\documentclass[letterpaper, 10 pt, conference]{ieeeconf}
\usepackage[english]{babel} 

\IEEEoverridecommandlockouts 

\usepackage{cite}

\usepackage[pdftex]{graphicx}
\usepackage[unicode]{hyperref}
\usepackage{subcaption}
\usepackage[export]{adjustbox}
\usepackage{varwidth}

\usepackage[usenames,dvipsnames,svgnames,table]{xcolor} 
\usepackage{amssymb}
\usepackage{amsmath}
\usepackage{txfonts}
\usepackage{times}
\usepackage{booktabs}
\newcommand{\ra}{\renewcommand{\arraystretch}{1.2}}

\renewcommand{\baselinestretch}{0.96}

\DeclareMathOperator*{\argmin}{arg\,min}
\DeclareMathOperator*{\argmax}{arg\,max}

\newcommand{\barg}{\textcolor{blue}{Thread of argument:}\begin{itemize}[label={\color{blue}\textbullet}]}

\newcommand{\earg}{\end{itemize}}

\DeclareTextCompositeCommand{\k}{LY1}{a}
{\oalign{a\crcr\noalign{\kern-.27ex}\hidewidth\char7}}
\DeclareTextCompositeCommand{\k}{LY1}{e}
{\oalign{e\crcr\noalign{\kern-.27ex}\hidewidth\char7\hidewidth}}
\DeclareTextCompositeCommand{\k}{LY1}{E}
{\oalign{E\crcr\hidewidth\char7\hidewidth}}

\usepackage{amsmath}
\usepackage{fancyhdr}

\usepackage{algorithm, algorithmic}

\usepackage{array}
\title{\LARGE \bf VALUE: Large Scale Voting-based Automatic Labelling \\for Urban Environments}

\author{Giacomo Dabisias$^{1,2}$, Emanuele Ruffaldi$^{1}$, Hugo Grimmett$^{2}$, and Peter Ondruska$^{2}$
	\thanks{$^{1}$PERCRO Laboratory, TeCIP Institute, Scuola Superiore Sant'Anna, Pisa, Italy.
		{\tt\small \{g.dabisias, e.ruffaldi\}@santannapisa.it}}%
	\thanks{$^{2}$Blue Vision Labs, London, United Kingdom. \newline
		{\tt\small \{giacomo, hugo, peter\}@bluevisionlabs.com}}%
}

\begin{document}
	\maketitle
	\thispagestyle{fancy}
	\pagestyle{fancy}



	\begin{abstract}
		This paper presents a simple and robust method for the automatic localisation of static 3D objects in large-scale urban environments. By exploiting the potential to merge a large volume of noisy but accurately localised 2D image data, we achieve superior performance in terms of both robustness and accuracy of the recovered 3D information. The method is based on a simple distributed voting schema which can be fully distributed and parallelised to scale to large-scale scenarios. To evaluate the method we collected city-scale data sets from New York City and San Francisco consisting of almost 400k images spanning the area of 40 km$^2$ and used it to accurately recover the 3D positions of traffic lights. We demonstrate a robust performance and also show that the solution improves in quality over time as the amount of data increases.
	\end{abstract}

	\section{Introduction}
\label{introduction}

\begin{figure}
	\includegraphics[width=\columnwidth, right]{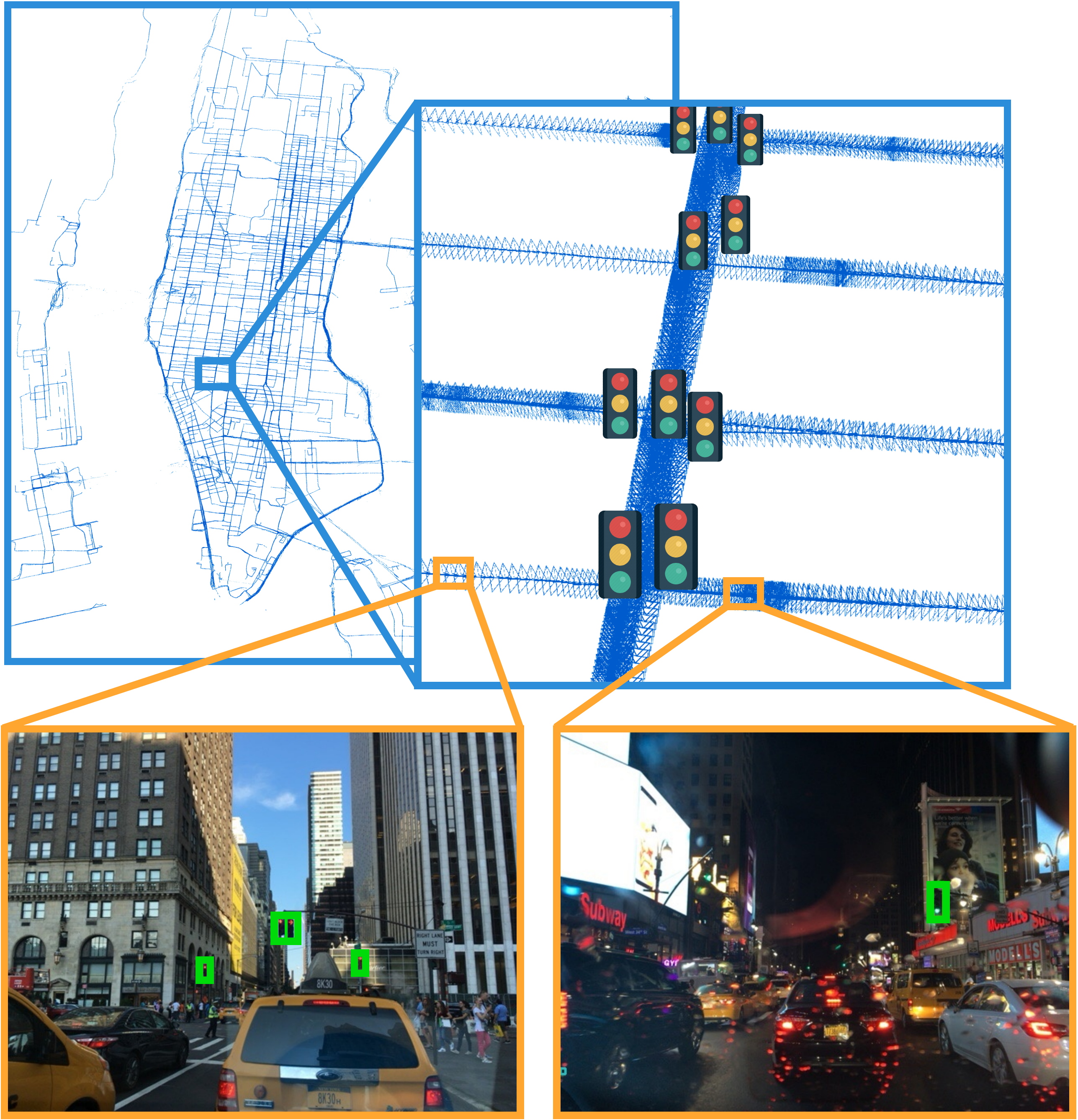}
	\caption{We use VALUE to automatically find the 3D locations of all traffic lights in Manhattan, such that they can be used for autonomous driving. Zero images were manually labeled to make this semantic map.}
	\label{fig:map}
\end{figure}

The next generation of self-driving cars will likely operate more robustly by using maps of their environment \cite{janai2017computer}. These maps allow the robots to have strong priors on their environments to improve perception \cite{barnes2015exploiting}, and have metric and semantic components for localisation and planning (\textit{top-left} and \textit{top-right} in Fig.~\ref{fig:map}) \cite{nuchter2008towards}. For self-driving cars in urban environments, these semantic maps typically contain static objects such as road signs, traffic lights, road markings, etc. It is common to manually label these \cite{stallkamp2011german}, but on the city-scale this becomes prohibitively expensive, and furthermore it needs to be laboriously relabelled as the urban landscape inevitably changes. A system that can automatically generate such labels for entire cities without the need for hand-labeling would be very valuable to overcome this issue. In this paper we present such a system.

While the idea of using machine learning for automatic content detection has been explored before, the use of large-scale accurate maps together with machine learning opens new possibilities. Our work is inspired by the possibility to merge large amounts of noisily labelled but accurately localised data from a particular location. With this, we compute accurate, denoised estimations of the semantic information of a superior quality.

Our contribution is a novel and simple system that (1) takes images, their accurate 3D position in a large-scale environment and returns the 3D positions of static object in the environment; (2) improves both object detection rate and 3D object position accuracy as areas are revisited, making it scale and work better with very large data sets; (3) deliberately avoids the need for visually matching objects in-between the images, a problem that prevents most similar systems from working on objects that (a) inherently look similar to each other (like traffic lights) and (b) can appear very differently based on time of day, lighting, weather conditions, season, etc.

We evaluate our system on two new large datasets taken from San Francisco and New York City, in total comprising almost 0.4M images over 40 $km^2$ from different times and weather conditions over a period of several weeks to robustly recover position of the traffic lights in the environment. We demonstrate that significantly superior results can be obtained using even only mediocre and noisy 2D detection algorithms, if enough data are provided.

	\section{Background} 
\label{Background}

A number of works explore robust detection of static
3D objects in the environment.

The basic component in a vision-based systems is an
accurate 2D detection of the object in a single image or video. Recently, this approach has been dominated by
deep learning techniques \cite{john2014traffic, csurka2008simple, pathak2014fully}. In this work we decided to adopt a fully Convolutional Neural Network (CNN) as described by Huang et al. \cite{HuangLW16a} to detect traffic lights in a large collection of pictures.

Given two detections of the same object coming from a stereo camera, it is possible to determine the 3D position of the object by triangulation \cite{hartley1997triangulation, lee2016critical}. These approaches are viable, for example, for the online detection of relative positions of objects around the vehicle

Similarly, if the position of multiple cameras observing the same object are known, a multi-view triangulation approach can be used \cite{hartley2003multiple}. This has an advantage over live detection since a significantly higher number of potential views from different cameras can be collected over extended periods of time, resulting into potentially significantly higher performance. We take this approach in our work when we use large collection of 2D observations to produce accurate, denoised 3D locations.

A common problem in applying such approaches relies in the need to accurately localise a set of sensors in the area. This problem can be addressed by using a highly-precise GPS system. Unfortunately, accuracy of GPS in dense urban environments is limited due to low sky visibility. An alternative approach in these environments consists in using a map-based localisation. Here state-of-the-art structure-from-motion systems have demonstrated ability to construct large-scale map of the environments \cite{klingner2013street}. We follow this approach and construct a large-scale map of the city to localise accurately the positions of all the traffic lights.

The most closely related work to ours is \cite{fairfield2011traffic} where they use, similarly to us, high-accuracy localisation of camera-equipped vehicles to map positions of traffic lights in the environment. The main difference relies in the fact that they use lidars for localisation, high resolution cameras for image acquisition and a traffic light position prior database. Our work is novel for the use of a robust triangulation method operating jointly on the set of all the data from a particular location, resulting in improved performance as the amount of data increases.

	\section{Method}
\label{Method}


Our system takes a large set of 2D images $I_i$, with associated camera-intrinsics parameters $q_i$ and 6 degrees-of-freedom poses $P_i \in \mathbb{SE}(3)$, and produces a set of 3D positions of objects $L_i \in \mathbb{R}^3$. In our work the images are captured from our mapping fleet traversing various cities, and the poses are calculated using a large-scale structure-from-motion (SFM) pipeline \cite{koenderink1991affine}, but in general there is no restriction on the source of the poses as long as they are accurate and globally consistent.

The process then consists of two steps: (1) applying a noisy 2D detector to each image $I_i$ resulting in a set of object detections $Z_i \subset \mathbb{R}^2$ followed by (2) estimating their final 3D positions $L$ by a simple voting-based triangulation algorithm.

\subsection{2D Object Detection}
\label{ssec:detection}

We generate 2D object detections in the images using an off-the-shelf CNN trained to predict bounding boxes for traffic lights \cite{HuangLW16a}. These detections are usually noisy and suffer from many false positives and false negatives. In Sec.~\ref{Results} we show that our system compensates for these noisy detections if shown a large amount of data. One alternative to using a detector is to use hand-annotated labels from, for example, \textit{Amazon Mechanical Turk} \cite{buhrmester2011amazon} which have also been shown to suffer from label noise \cite{frenay2014classification}.

\begin{algorithm}[t]
\caption{Robust Voting-Based Triangulation}
\begin{tabular}{l l l}
Input: & $\mathbf{I}$ & \textit{set of images}\\
          & $\mathbf{Q}$ & \textit{camera intrinsics}\\
          & $\mathbf{P}$ & $\mathbb{SE}(3)$ \textit{camera poses}\\
          & $d_{max}$ & \textit{maximum reprojection error}\\
          & $\alpha$ & \textit{minimum ratio of inliers}\\
Output: & $\mathbf{L}$ & \textit{3D positions of objects}
\vspace{1mm}
\end{tabular}
\begin{tabular}{r l}
\vspace{1mm}
& Detect objects in 2D images:\\
1. & \textbf{for} $I_i \in \mathbf{I}$\\
2. & \hspace{4mm} $Z_i \leftarrow \textit{detect}(I_i)$\\ 
3. & $\mathbf{Z} \leftarrow \cup_i \hspace{1mm} Z_i$\\
4. & $\mathbf{L} \leftarrow \emptyset$\\
5. & \textbf{for} $(i_a,i_b) \in \mathbf{I}$\\
6. & \hspace{2mm} \textbf{for} $(z_a,z_b) \in {(\mathbf{Z}_{i_a},\mathbf{Z}_{i_b)}}^2$\\
\vspace{1mm}
& \hspace{4mm} Compute 3D position of the object:\\
7. & \hspace{4mm} $l_{ab} \leftarrow \textit{triangulate}(\{z_a,z_b\})$\\
\vspace{1mm}
& \hspace{4mm} Compute inliers for computed 3D position:\\
8. & \hspace{4mm} $S_{ab} \leftarrow \{ z_k | \forall z_k \in \mathbf{Z} : \pi(l_{ij}, p_k, q_k) - z_k < d_{max} \}$\\
\vspace{1mm}
& Find the hypothesis with most votes:\\
9. & \hspace{4mm} $a,b \leftarrow \argmax_{a,b} |S_{ab}|$\\
10. & \textbf{if} $|S_{ab}| \geq \alpha \cdot mean(|S|)$\\
11. & \hspace{4mm} $L \leftarrow L \cup \textit{triangulate}(S_{ab})$\\
12. & \hspace{4mm} $Z \leftarrow Z - S_{ab}$\\
13. & \hspace{4mm} \textbf{goto} $5$\\
14. & \textbf{return} $\mathbf{L}$
\end{tabular}
\label{algo:triangulation}
\end{algorithm}

\subsection{Robust Voting-based Triangulation}
\label{ssec:clustering}

The output of the previous step is a large set of 2D detections. Importantly, the 2D detection step cannot tell you which detections can be associated with which physical 3D traffic light $D_i$, and any feature descriptors that it might produce to associate them would be useless under the appearance changes that we see in outdoor environments. This is true for any set of objects that look similar (traffic lights are a good example). The only difference between them is their position in 3D space. Without this association, classical algorithms for multi-view triangulation can therefore not be directly used. Instead, we use a robust voting-based triangulation algorithm to jointly determine these 2D associations and the position of the traffic lights in 3D space.

\begin{figure*}
	\centering
	\includegraphics[width=\textwidth]{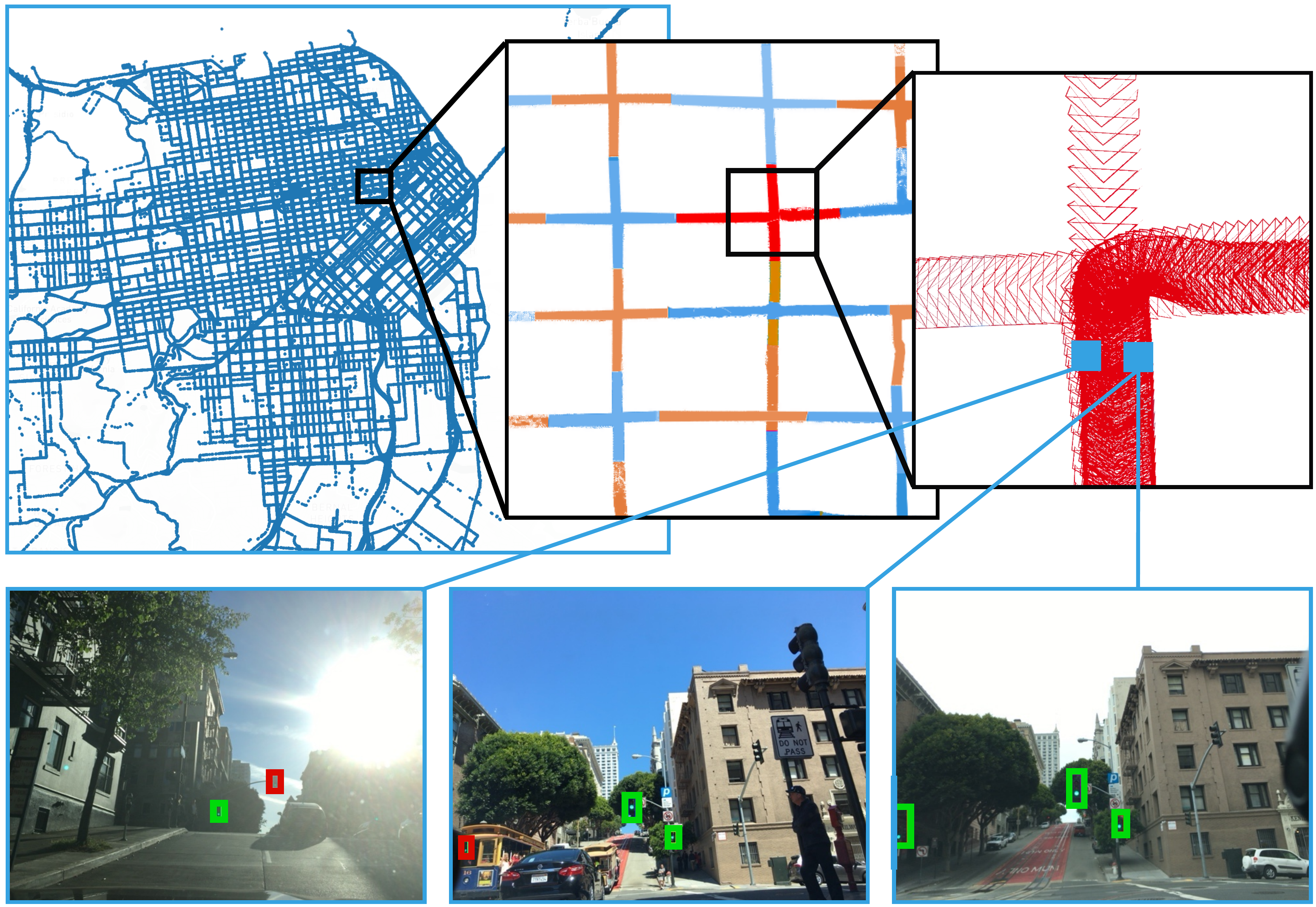}
	\caption{An example of a cluster in the map consisting of several traversals in differing conditions. \textit{Top}: Close-up of the dataset clustering. \textit{Bottom row}: example frames that are from different traversals belonging to the highlighted cluster and associated traffic lights.}
	\label{fig:general}. 
\end{figure*}

For each pair of detections $(z_a, z_b)$ (where $a$ and $b$ are indices into 2D detections) from two different images $I_i$, $I_j$ we create a 3D hypothesis $h_{ab}$ under the assumption that these two detections correspond to the same physical 3D traffic light generating in total $\mathcal{O}(N^2)$ hypotheses where $N$ is the total number of detected traffic lights. The 3D position $l^*$ of each hypothesis can be determined by $K$-view triangulation (in this case $K=2$), where we minimise the sum of the reprojection errors:
\begin{equation}
l^* = \argmin_l \sum_{k\in W} \big(\pi(l, p_k, q_k) - z_k \big)^2,
\label{eq:reprojection}
\end{equation}
where $W$ is \{a,b\} in this case, $\pi$ is the projection of the 3D point $l$ into the camera at position $p_k$ with intrinsics $q_k$ and $z_k$ is the center of the closes traffic light in                            the image. We consider a hypothesis viable if it satisfies the following:
\begin{enumerate}
\item \emph{triangulation constraint}: the point is triangulated in front of each camera in a valid distance range,
\item \emph{rays intersect in 3D space}: the reprojection error is smaller than $d_{\text{max}}$,
\item \emph{the projection is stable}: the angle between the optical axes is larger than $\theta_{min}$,
\item \emph{distance to camera}: the distance from the traffic light to either camera is less than $r_{max}$.
\end{enumerate}
Optionally, additional constraints reflecting prior information about the location of a traffic lights can be used to further restrict the hypothesis space.

For each hypothesis $h_{ab}$ we compute the set of consistent inliers $S_{ab}$. This set consists of all the 2D detections that observe a traffic light at the same location, which is computed by projecting the 3D position $l^*$ into each image and verifying whether the projected position is less than $d_{\text{max}}$ to any 2D detection. This parameter can be tuned to obtain a lower reprojection error at the cost of the number of correctly triangulated traffic lights. Next, we remove the hypothesis with the maximum number of votes and also remove the detections that voted for it (inlier detections). This process is repeated until no hypothesis with at least $\alpha \cdot M$ inliers is found, where $M$ is the average number of inliers per hypothesis and $\alpha$ is a tunable parameter over the confidence. This creates a set of confirmed hypotheses. An important theoretical property of this schema is that in the case of noisy but unbiased 2D detector and a uniform distribution of the data, it converges to the correct solution as the amount of data increases. This is due to noisy detections forming hypotheses with small numbers of votes, and correct detections gathering consistent votes over time. As the amount of data increases, these two metrics begin to separate, and $\alpha$ is the threshold on their ratio.

Finally, for every hypothesis we refine its 3D position by optimising the reprojection error from \eqref{eq:reprojection} over all the hypothesis detections. This entire algorithm is presented in Alg.~\ref{algo:triangulation}.

\subsection{Large-Scale System}
\label{ssec:lss}

The above method works well for small-scale scenario but does not scale well to large, city-scale settings we are interested in due to its potential $\mathcal{O}(N^3 * M^2)$ complexity where $M$ is the number of images $N$ is the average number of detected traffic lights in each image. Instead, we resort to a distribution schema based on splitting the data set to clusters, running Alg.~\ref{algo:triangulation} for each cluster independently, and then merging the results.

We employ a simple clustering schema where, for each traffic light, we add to its cluster all images closer than $N_{\text{max}}$ meters (50m default). An illustration of these clusters is shown in Fig.~\ref{fig:general}.

After each cluster is triangulated using Alg.~\ref{algo:triangulation}, it might be the case that the same traffic light is triangulated in two different clusters. To resolve this issue we merge all pairs of traffic lights closer than $1 m$, producing the final set of labels $L$. It would be possible to further optimize the final results by globally optimizing the positions of the traffic lights in the map, but this has not been done given the high computational cost. 

	\section{Experiments}
\label{Results}

We evaluate the presented system on two large-scale data sets from San Francisco and New York City that we collected using a dedicated fleet of mapping vehicles. We demonstrate that the presented system scales to the size of cities, and that as the amount of data increases it generates increasingly accurate results both in terms of successfully recovered traffic lights and their 3D positions, despite using a very inaccurate off-the-shelf 2D detector.

\begin{table}[b]
	\ra
	\centering
	\begin{tabular}{@{}l r r@{}}
		\hline\toprule 
		 & \emph{San Francisco} & \emph{New York City}  \\\midrule
		\# images & 12048 & 360207 \\ \hline
		\# detections & 17198 & 547689 \\ \hline
		\# clusters & 172 & 3941 \\ \hline
        \# traffic lights & 183 & 1732 \\ \hline
        \# detectable traffic lights & 167 & 1906 \\ \hline
		mean \# views / cluster& 70.05 &  91.4 \\ \hline
		mean \# detections / frame & 1.65&   2.84 \\ \hline
        \# images with 0 detection & 1587 & 44483 \\ \hline
		\# images with 1 detections & 6231 & 145608 \\ \hline
		\# images with 2 detections & 2389 & 125924 \\ \hline
        \# images with 3 detections & 1346 & 32487 \\ \hline
        \# images with 4 detections & 369 & 7754 \\ \hline
        \# images with 5 detections & 98 & 2700 \\ \hline
        \# images with 6 detections & 17 & 802 \\ \hline
        \# images with 7 detections & 9 & 283 \\ \hline
        \# images with 8 detections & 0 & 80 \\ \hline
        \# images with 9 detections & 1 & 51 \\ \hline
        \# images with 10+ detections & 1 & 35 \\ 
		\bottomrule \hline
	\end{tabular}
	\caption{Per-dataset statistics of 2D detection and clustering.}
	\label{table:datasets}
\end{table}



\begin{table}[t]
\centering
\ra
\begin{tabular}{@{}lrr@{}}
\hline\toprule
\textbf{} & \emph{San Francisco} & \emph{New York City}  \\\midrule
true positives & 156 & 1560 \\ \hline
false positives & 4 &  84 \\ \hline
false negatives & 11 &  172 \\ \hline
duplicates & 14&      56 \\ \hline
mean reprojection error & 2.94&  3.24 \\ 
\bottomrule\hline
\end{tabular}
\caption{The results of the method on two datasets.}
\label{triangulationStatistics}
\end{table}

\subsection{Data Sets}
The San Francisco and New York City data sets have been created by capturing images using a fleet of vehicles. The vehicles traversed most of the roads multiple times, in both directions, at varying times of day and weather conditions. During this time they captured images at regular intervals. Example images are shown in Figs.~\ref{fig:map} and \ref{fig:general}. Each of these images has associated ground-truth 2D labels of traffic lights.

We resize each image to $640\times480$ pixels and use a large-scale, distributed, structure-from-motion pipeline \cite{klingner2013street, zhuparallel} running on multiple computers to calculate the 3D positions $P$ of the images. 

Each data set covers an area with a certain number of physical traffic lights. Not all of them are recoverable, i.e. their 3D positions cannot be accurately determined. We consider a traffic light recoverable if it has been observed from at least two different viewpoints under angle difference at least $\theta_{min}$. In reality, as the amount of data increases, almost all the traffic lights eventually become recoverable. The sizes of these data sets together with their RMSE results are shown in Tab.~\ref{table:datasets}. We also present the amount of traffic lights present in our dataset along with the number of detectable traffic lights.  
\subsection{2D Detection}
We use a simple, convolutional neural network architecture to detect traffic lights in 2D images. Firstly, we use a binary segmentation network \cite{HuangLW16a} to compute the probability that each pixel is part of a traffic light. We then use a simple thresholding schema to compute connected components of pixels representing traffic lights, and fit bounding boxes.

We train this network using the Bosch Data Set \cite{BehrendtNovak2017ICRA}. We split the data set into a training set of size 5,093 and a testing set of size 8,334, covering in total 24,242 traffic lights detections. For training we use two Nvidia P5000 cards until convergence. The network has been used to classify all the images in our datasets.

In this work we purposely did not fine-tune the detector for either the San Francisco or New York City data sets. There are significant differences between the training and testing data: the Bosch data are from a suburban area, while our data are urban; the cameras are different; and the training data contain mostly small traffic lights while our evaluation data contain traffic lights of all sizes. While the learned classifier achieves 90\% recall on the Bosch test data, it becomes a relatively noisy detector on our datasets with a recall of 85\% and an average Intersection-Over-Union (IOU) of 0.45. Fine tuning the detector could possibly improve further the results, but this has not been investigated.


\subsection{Results}

\newcommand{\figsize}{0.40\linewidth}
\begin{figure*}
 \centering
 \begin{subfigure}[b]{\figsize}
	\includegraphics[width=\linewidth]{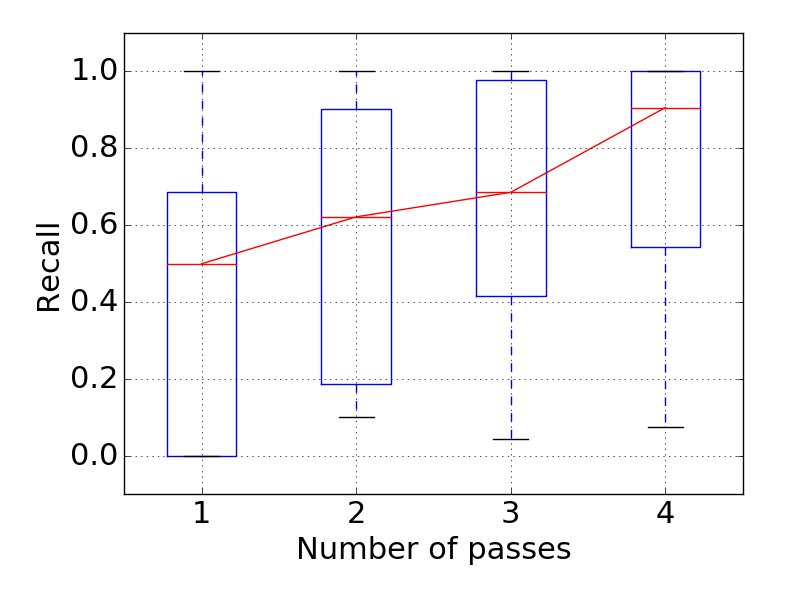}
 	\caption{New York City}
 	\label{fig:pdetectionNYC}
 \end{subfigure}
 \begin{subfigure}[b]{\figsize}
	\includegraphics[width=\linewidth]{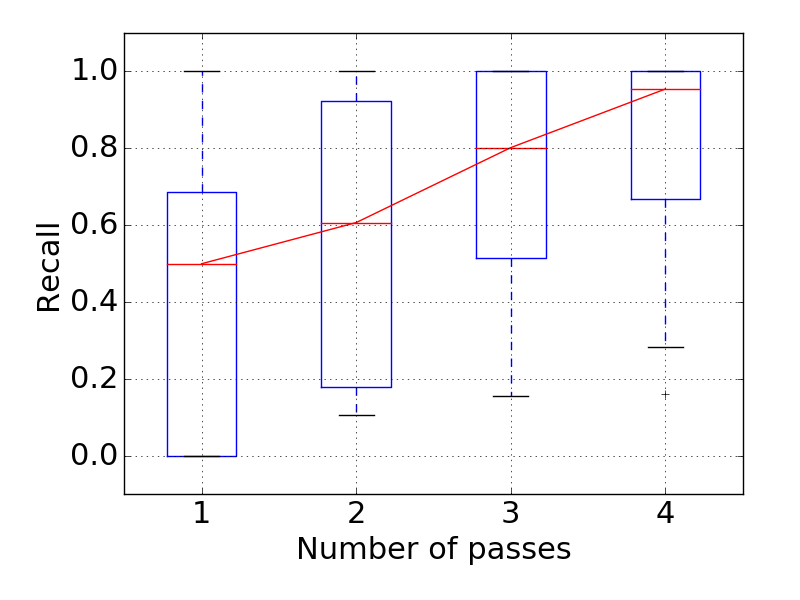}
	\caption{San Francisco}
	\label{fig:pdetectionSF}
\end{subfigure}
\begin{subfigure}[b]{\figsize}
	\includegraphics[width=\linewidth]{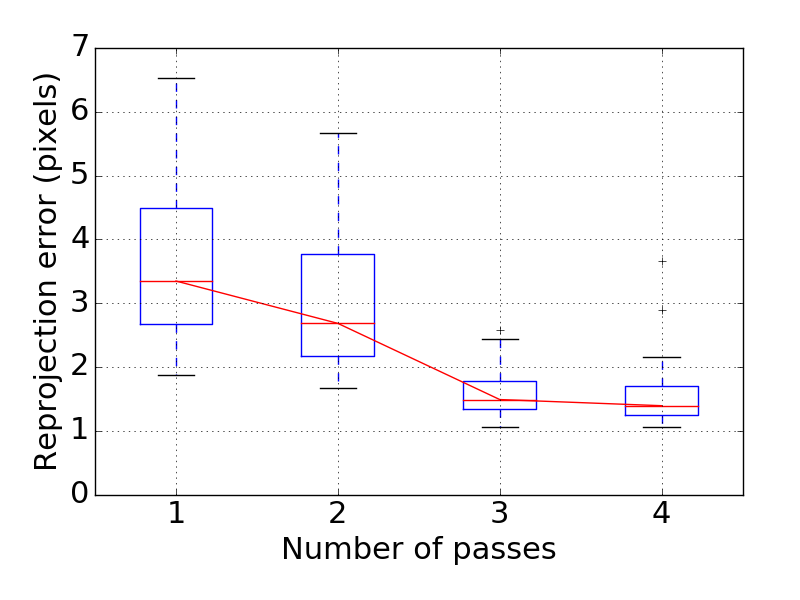}
	\caption{New York City}
	\label{fig:reNYC}
\end{subfigure}
\begin{subfigure}[b]{\figsize}
	\includegraphics[width=\linewidth]{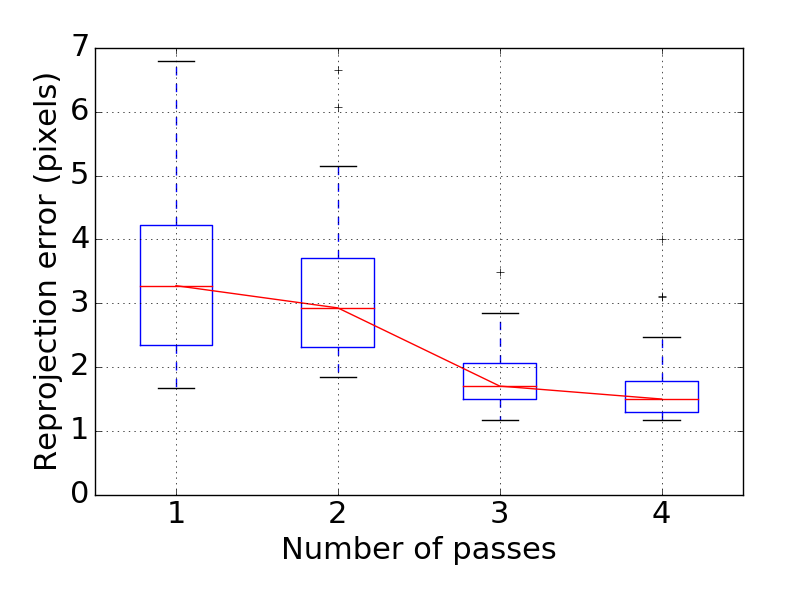}
	\caption{San Francisco}
	\label{fig:reSF}
\end{subfigure}
 \caption{Performance of the system as a function of number of passes through a location. As the amount of data increases both recall and 3D localisation accuracy (as measured by negative reprojection error) increase.}
 \label{fig:PDetectionAndNPasses}
\end{figure*}

We run the described clustering method from Section \ref{Method} on all the data. Fig.~\ref{fig:general} shows the results of clustering in the San Francisco data set. Any clusters can contains images from multiple passes of the mapping vehicles through the area, as shown in Tab.~\ref{tab:NPassesInClusters}. 

\begin{table}[]
	\centering
	\ra
	\begin{tabular}{@{}cc c@{}}
		\hline\toprule
		&            \multicolumn{2}{c}{\emph{\# clusters}} \\
		\emph{\# passes} & \emph{San Francisco} &  \emph{New York City} \\
		\midrule
		1 & 47 & 476   \\ \hline
		2 & 43 &  304   \\ \hline
		3 & 9  &  229   \\ \hline
		4 & 6  &  121   \\ \hline
		5 & 1  &  49   \\ \hline
		6 & 4  &  27  \\ \hline
		7 & 6 &   0 \\ \hline
		8 & 6 & 0  \\ \hline
		9 & 1 &    0  \\
		\bottomrule\hline
	\end{tabular}
	\caption{Statistics of a number of passes per cluster. As the mapping fleet traverses the environment each place is visited several times. As discussed in the text more passes through the environment result into higher performance of the system.}
	\label{tab:NPassesInClusters}
\end{table}

Tab.~\ref{triangulationStatistics} shows the statistics for the triangulation step. The presented method is able to recover at least 90\% of all the recoverable traffic lights in both data sets, while suffering from only about 10\% of false positive detections. The average reprojection error of the triangulated objects for two datasets is 2.94 and 3.24 pixels for San Francisco and New York respectively. Note that reprojection error incorporates both the error in the triangulated 3D object and the underlying map accuracy. As discussed in Sec.~\ref{ssec:lss}, during the triangulation some of the traffic lights might be detected in two or more different clusters and must be unified in the merge step. These form only a small fraction of all the traffic lights. All results have been inspected manually given the absence of ground truth for the 3D positions of traffic lights.

\renewcommand{\figsize}{0.32\linewidth}
\begin{figure*}[ht]
	\centering
	\begin{subfigure}[b]{\figsize}
		\includegraphics[width=\linewidth]{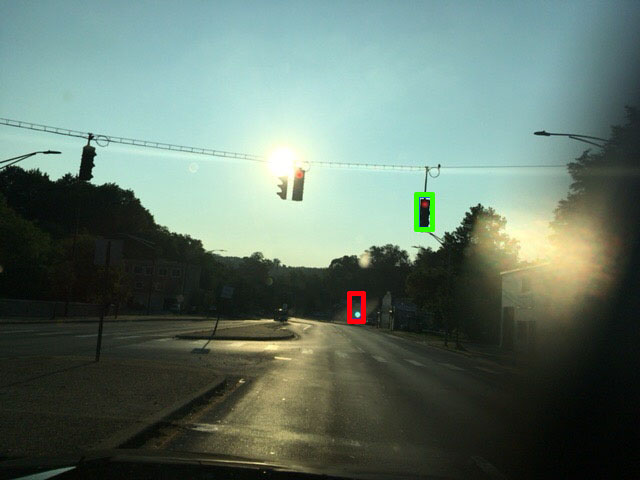}
		\caption{}
		\label{fig:failure1}
	\end{subfigure}
	\begin{subfigure}[b]{\figsize}
		\includegraphics[width=\linewidth]{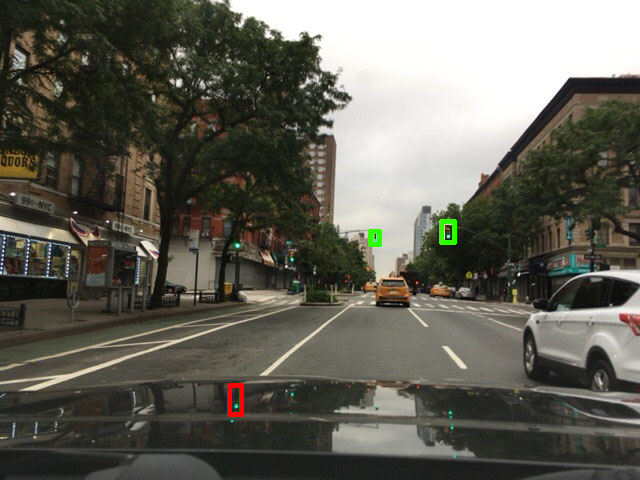}
		\caption{}
		\label{fig:failure2}
	\end{subfigure}
	\begin{subfigure}[b]{\figsize}
		\includegraphics[width=\linewidth]{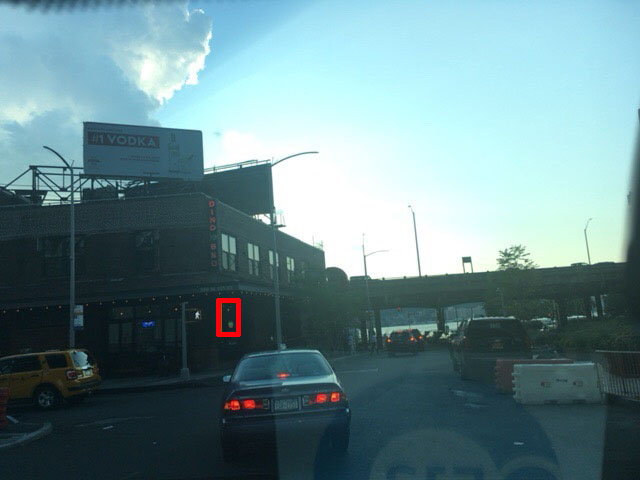}
		\caption{}
		\label{fig:failure3}
	\end{subfigure}
	\caption{Failure modes of the presented method. When not enough data from a particular location are provided both (a) false negative detections due to the undetected objects or (b) temporarily consistent wrong detections can manifest. The largest challenge is (c) consistent and repeated detections of a objects that look similar to a traffic light over a period of time.}
	\label{fig:failuremodes}
\end{figure*}

Running the 2D detector over all 372k images took 30 hours. Clustering was performed in 1.1 minutes for San Francisco and 19 minutes for New York City, while the final triangulation took 1.1 and 3.1 minutes respectively.


Of primary importance is the performance of the system as the number of data increases. We characterise this by an increasing number of passes through an area. For this experiment we took a random subset of 25 clusters with at least 5 passes and computed the statistics of the number of recovered traffic lights and their reprojection error. The results are shown in Fig.~\ref{fig:PDetectionAndNPasses}(a-b). Note that the recall starts off poorly because not all traffic lights in a cluster are detectable in a single pass, if for instance, they are angled orthogonally to the direction of travel and increases with the number of passes through an area. With increasing passes we are more likely to detect the traffic light and have enough views to accurately localise it in 3D. We can see that the likelihood of detecting a traffic light does increase with number of passes. Note that the number of false positives for this random subset of 25 clusters is zero.

Figs.~\ref{fig:PDetectionAndNPasses}(c-d) show that the 3D localisation accuracy also increases with number of passes. We measure this by taking the 3D object position estimated using up to $N$ passes and project it into the images from a \emph{leave-one-out} pass, measuring the reprojection error between where that 3D object is projected into the image and where it truly appears in the image. The statistics in Fig.~\ref{fig:PDetectionAndNPasses} show that the system converges in both recoverability of the traffic lights and their 3D position with more data.

Finally, we show some failure cases. The method converges under the assumption that the underlying detector is unbiased, but might produce incorrect results when this assumption is broken or when only a small amount of data are provided. Figure \ref{fig:failuremodes} shows some of these cases. While the method is prone to generate a number of false negative detections, the largest problem are the false positive detections created by consistent, incorrect detections. While some of them tend to appear only in a single pass (such as series of reflections on a deck of the car) and thus can be resolved with more data, the incorrect detections of traffic-light-like objects which are repeatedly and consistently observed over periods of time poses a serious challenge to be addressed in the future.

	\section{Conclusions}
\label{Conclusions}

We have demonstrated a simple and robust system for finding the 3D positions of static objects in complex city environments. We leverage a reliable image pose source and a large quantity of image data to overcome the common challenges of noisy 2D labels. The resulting accurate 3D object positions are borne out of a voting-based triangulation system that solves the data association problem that poses a particularly difficult challenge when the desired objects are similar in physical construction and yet appear vastly different in images as a result of strong variations in lighting and weather. 

The system is specifically designed in such a way to be parallelisable, and therefore efficiently process very large image sets. We have evaluated our system on city-scale data sets comprising almost 0.4M images, and have shown that despite the very noisy input detections, the system output increases in 3D positional accuracy and recall with more data.

	\bibliographystyle{IEEEtran}
	\bibliography{root}

\begin{thebibliography}{10}
\providecommand{\url}[1]{#1}
\csname url@rmstyle\endcsname
\providecommand{\newblock}{\relax}
\providecommand{\bibinfo}[2]{#2}
\providecommand\BIBentrySTDinterwordspacing{\spaceskip=0pt\relax}
\providecommand\BIBentryALTinterwordstretchfactor{4}
\providecommand\BIBentryALTinterwordspacing{\spaceskip=\fontdimen2\font plus
\BIBentryALTinterwordstretchfactor\fontdimen3\font minus
  \fontdimen4\font\relax}
\providecommand\BIBforeignlanguage[2]{{%
\expandafter\ifx\csname l@#1\endcsname\relax
\typeout{** WARNING: IEEEtran.bst: No hyphenation pattern has been}%
\typeout{** loaded for the language `#1'. Using the pattern for}%
\typeout{** the default language instead.}%
\else
\language=\csname l@#1\endcsname
\fi
#2}}

\bibitem{janai2017computer}
J.~Janai, F.~G{\"u}ney, A.~Behl, and A.~Geiger, ``Computer vision for
  autonomous vehicles: Problems, datasets and state-of-the-art,'' \emph{arXiv
  preprint 1704.05519}, 2017.

\bibitem{barnes2015exploiting}
D.~Barnes, W.~Maddern, and I.~Posner, ``Exploiting 3d semantic scene priors for
  online traffic light interpretation,'' in \emph{Intelligent Vehicles
  Symposium (IV), 2015 IEEE}.\hskip 1em plus 0.5em minus 0.4em\relax IEEE,
  2015, pp. 573--578.

\bibitem{nuchter2008towards}
A.~N{\"u}chter and J.~Hertzberg, ``Towards semantic maps for mobile robots,''
  \emph{Robotics and Autonomous Systems}, vol.~56, no.~11, pp. 915--926, 2008.

\bibitem{stallkamp2011german}
J.~Stallkamp, M.~Schlipsing, J.~Salmen, and C.~Igel, ``The german traffic sign
  recognition benchmark: a multi-class classification competition,'' in
  \emph{Neural Networks (IJCNN), The 2011 International Joint Conference
  on}.\hskip 1em plus 0.5em minus 0.4em\relax IEEE, 2011, pp. 1453--1460.

\bibitem{john2014traffic}
V.~John, K.~Yoneda, B.~Qi, Z.~Liu, and S.~Mita, ``Traffic light recognition in
  varying illumination using deep learning and saliency map,'' in
  \emph{Intelligent Transportation Systems (ITSC)}.\hskip 1em plus 0.5em minus
  0.4em\relax IEEE, 2014, pp. 2286--2291.

\bibitem{csurka2008simple}
G.~Csurka and F.~Perronnin, ``A simple high performance approach to semantic
  segmentation.'' in \emph{BMVC}, 2008, pp. 1--10.

\bibitem{pathak2014fully}
D.~Pathak, E.~Shelhamer, J.~Long, and T.~Darrell, ``Fully convolutional
  multi-class multiple instance learning,'' \emph{arXiv preprint
  arXiv:1412.7144}, 2014.

\bibitem{HuangLW16a}
G.~Huang, Z.~Liu, and K.~Q. Weinberger, ``Densely connected convolutional
  networks,'' \emph{CoRR}, vol. abs/1608.06993, 2016.

\bibitem{hartley1997triangulation}
R.~I. Hartley and P.~Sturm, ``Triangulation,'' \emph{Computer vision and image
  understanding}, vol.~68, no.~2, pp. 146--157, 1997.

\bibitem{lee2016critical}
H.-L. Lee, ``Critical points for two-view triangulation,'' \emph{arXiv preprint
  1608.05512}, 2016.

\bibitem{hartley2003multiple}
R.~Hartley and A.~Zisserman, \emph{"Multiple view geometry in computer
  vision"}.\hskip 1em plus 0.5em minus 0.4em\relax Cambridge university press,
  2003.

\bibitem{klingner2013street}
B.~Klingner, D.~Martin, and J.~Roseborough, ``Street view
  motion-from-structure-from-motion,'' in \emph{Proceedings of the IEEE
  International Conference on Computer Vision}, 2013, pp. 953--960.

\bibitem{fairfield2011traffic}
N.~Fairfield and C.~Urmson, ``Traffic light mapping and detection,'' in
  \emph{ICRA}.\hskip 1em plus 0.5em minus 0.4em\relax IEEE, 2011, pp.
  5421--5426.

\bibitem{koenderink1991affine}
J.~J. Koenderink and A.~J. Van~Doorn, ``Affine structure from motion,''
  \emph{JOSA A}, vol.~8, no.~2, pp. 377--385, 1991.

\bibitem{buhrmester2011amazon}
M.~Buhrmester, T.~Kwang, and S.~D. Gosling, ``Amazon's mechanical turk: A new
  source of inexpensive, yet high-quality, data?'' \emph{Perspectives on
  psychological science}, vol.~6, no.~1, pp. 3--5, 2011.

\bibitem{frenay2014classification}
B.~Fr{\'e}nay and M.~Verleysen, ``Classification in the presence of label
  noise: a survey,'' \emph{IEEE transactions on neural networks and learning
  systems}, vol.~25, no.~5, pp. 845--869, 2014.

\bibitem{zhuparallel}
S.~Zhu, T.~Shen, L.~Zhou, R.~Zhang, J.~Wang, T.~Fang, and L.~Quan, ``Parallel
  structure from motion from local increment to global averaging,'' \emph{ArXiv
  e-prints}, 2017.

\bibitem{BehrendtNovak2017ICRA}
K.~Behrendt and L.~Novak, ``A deep learning approach to traffic lights:
  Detection, tracking, and classification,'' in \emph{ICRA}.\hskip 1em plus
  0.5em minus 0.4em\relax IEEE, 2017, pp. 1370--1377.

\end{thebibliography}

\end{document}